# How important is language for human-like intelligence?

Gary Lupyan[1]; Hunter Gentry[2]; Martin Zettersten[3]

1. Department of Psychology, University of Wisconsin-Madison
2. Department of Philosophy; Cognitive Science Program, University of Central Florida
3. Department of Cognitive Science, University of California-San Diego

Abstract

We use language to communicate our thoughts. But is language merely the expression of thoughts, which are themselves produced by other, nonlinguistic parts of our minds? Or does language play a more transformative role in human cognition, allowing us to have thoughts that we otherwise could (or would) not have? Recent developments in artificial intelligence (AI) and cognitive science have reinvigorated this old question. We argue that language may hold the key to the emergence of both more general AI systems and central aspects of human intelligence. We highlight two related properties of language that make it such a powerful tool for developing domain-general abilities. First, language offers compact representations that make it easier to represent and reason about many abstract concepts (e.g., exact numerosity). Second, these compressed representations are the iterated output of collective minds. In learning a language, we learn a treasure trove of culturally evolved abstractions. Taken together, these properties mean that a sufficiently powerful learning system exposed to language—whether biological or artificial—learns a compressed model of the world, reverse engineering many of the conceptual and causal structures that support human (and human-like) thought.

Language is clearly an important source of knowledge about the world. It is largely because of language that our knowledge far exceeds our personal experiences. Language allows us to learn about past events, helps to plan for the future, and is indispensable for creating the stories that comprise human culture. Language is also at the center of much of our formal education. Given its seeming importance, it is therefore surprising that the role of linguistic input has figured minimally in accounts of how humans acquire and structure semantic knowledge (Barsalou, 1999; Tulving, 1972)[1]. In explaining the origins and development of semantic knowledge, the more empiricist cognitive scientists have tended to focus on the role of perception and sensorimotor grounding (Prinz, 2004; Rogers & McClelland, 2004). Such work acknowledges that we might ordinarily learn many facts about water, such as its chemical structure, from language, but the more important conceptual "cores" are learned from our interactions with the world. We don't need language to learn that water is wet. On the more rationalist side, theorists have instead stressed the importance of innate conceptual knowledge and abstract reasoning (Bedny & Saxe, 2012; Quilty-Dunn et al., 2022). In AI, mastery of natural language has been a longstanding goal: a system that could understand natural language could be controlled by being spoken to. The open-endedness of language also meant that its successful use by a machine could be used as an intelligence metric, as in Turing's famous imitation game. The use of language was thus viewed as a problem to be solved by a sufficiently intelligent system rather than a way of making a system appropriately intelligent.

But what if this thinking is backwards? Could language be a key ingredient in creating human-like intelligence in the first place? Striking evidence for this possibility comes from what happens when artificial neural networks are exposed to large amounts of language. Large language models (LLMs) like ChatGPT, Claude, and Gemini are neural network transformers trained through self-supervised prediction of upcoming text[2]. Given a context (a chunk of text), the model makes predictions of what is likely to follow, adjusting its weights to minimize the error between its guess and what actually happens next. To learn that "Good" is frequently followed by "morning" does not require learning much beyond simple transition probabilities. But when a sufficiently large model is trained on a sufficiently large corpus, three surprising things happen.

First, the models learn language. While it was possible a few years ago to deny that LLMs "really" learn language (e.g., Dentella et al., 2023), it is no longer possible to reliably distinguish language produced by people and state-of-the-art LLMs (Hu et al., 2024; Sadasivan et al., 2025)[3]. A key innovation that led to this progress has been the development of the multi-head "attention" mechanism which allows the models to incorporate context into the internal representation of the linguistic prompt in a more powerful way than previously possible.[4] These models are generalized pattern learners. They learn language despite lacking specialized language-learning machinery of the sort that has long been argued to be necessary for any system to learn language (e.g., Lidz & Gagliardi, 2015). The same transformer architecture that powers modern language models can be put to use when performing image classification (Dosovitskiy et al., 2021) and to simulate chick development (Wood et al., 2024).

---

[1] Exceptions include neo-Whorfian work in the tradition of Melissa Bowerman and Stephen Levinson, work in developmental psychology including Sandy Waxman and Dedre Gentner, and Paul Harris et al's work on learning from testimony.

[2] Owing to lack of space, we are deliberately simplifying the training procedure. The bulk of training is indeed self-supervised prediction of text. However, the ability of LLMs to perform the many downstream tasks they are capable of requires a relatively small amount of supervised training to provide illustrations of, for example, what it means to summarize a document. Importantly, the effectiveness of this supervised training depends on having a sufficiently large base model trained through self-supervised language prediction.

[3] The impossibility of distinguishing whether a particular text was written by a person or an LLM does not entail that human and LLM-produced language is identical in aggregate. For example, there is compelling evidence that language produced by LLMs is more uniform compared language produced by people (Sourati et al., 2025).

Second, in the course of learning to produce human-like language, LLMs learn some of the very things that have been considered to be pre-requisites to language learning. For example, we know that human language understanding requires a heavy dose of pragmatic inference (Heintz & Scott-Phillips, 2023). Learning and using language also seems to require a system to be systematic: understanding "Mary loves John" implies that the system can understand "John loves Mary". But before being trained on language, LLMs know nothing of pragmatic inference and are hardly systematic. These abilities emerge in LLMs with exposure to language (Hu et al., 2023; Lepori et al., 2023).

Third, and perhaps most surprising of all, training these general-purpose networks on large amounts of natural language has enabled LLMs to perform a huge range of practical downstream tasks, ranging from summarizing and editing texts to diagnosing patients with apparently superhuman accuracy (Goh et al., 2024)[5]. The rapid uptake of LLMs across a wide swath of society speaks to the practical usefulness of these systems.

The wide-ranging abilities of LLMs raise two questions. First, is it a coincidence that these advances have come from using *natural language* as the main training data, or is there something special about language? Second, if language provides such effective training for artificial neural networks, might language input be more instrumental to *human* intelligence than many cognitive scientists have tended to assume (see Chalmers, 2024; Clatterbuck, 2024; Rothschild, forthcoming for recent philosophical treatments)?

The most direct way of answering the first question would require systematically comparing neural networks trained on language to those trained on only nonlinguistic data. If language is not necessary, it should be possible for nonlinguistic models to achieve the same performance (on nonlinguistic versions of tasks) as their language-trained counterparts. Nonlinguistic input should be sufficient (indeed in most cases more effective than language) for giving rise to the kinds of intelligence we find in non-human animals. After all, other animals manage to get by without the benefit of language. But when it comes to the types of intelligence that are more uniquely human–including a sophisticated theory of mind, relational and analogical reasoning, and the ability to learn a wide set of non-survival related skills–we predict that such nonlinguistic AI systems would struggle.

---

[4] We put scare-quotes around attention to highlight that it is misleading to think of this mechanism in terms of human attention. It is, more accurately, a form of adaptive kernel smoothing.

[5] Although it is difficult to compare data efficiency in an apples-to-apples way, it is certainly the case that LLMs are exposed to orders of magnitude more language than any person. Compared to biological systems, these models also require vast amounts of power to operate. These differences should temper the urge to draw overly strict analogies between artificial and biological neural networks.

According to some recent work in cognitive neuroscience, the answer to the second question is that no matter the usefulness of language for training AI models, its function in human cognition is highly circumscribed. For example, Fedorenko and colleagues have shown that explicit language tasks such as hearing or reading sentences activates a "language network" (which includes the left inferior frontal and middle frontal gyri, and anterior temporal lobe). This language network is not activated by nonverbal tasks such as numerical cognition, understanding actions, and tasks probing theory of mind (Fedorenko, Ivanova, et al., 2024). Based on this dissociation between brain networks involved in overtly linguistic tasks and in other cognitive tasks, Fedorenko et al. have argued that the role of language is strictly communicative and that thought is independent of language (Fedorenko, Piantadosi, et al., 2024). But although this work has been useful in helping us understand the neural substrates of language processing, it conflicts with a range of findings from cognitive science.

Just as manipulating linguistic experience is the best way to understand the role of language in artificial systems, we can ask if manipulating linguistic experience in humans affects human cognition. While we cannot deliberately deprive people of language, we can glean valuable insights from cases where people do not receive typical language input or from individuals who suffer from language impairments. We can also investigate typical development and study the associations between children's emerging language abilities to their cognition. Finally, we can experimentally manipulate the availability of language during a task and measure the effects of this manipulation on behavior.

These approaches, when taken together, provide converging evidence that language plays a transformative role in human cognition. For example, children who are born deaf and are not exposed to a conventional sign language struggle with a range of cognitive tasks such as theory of mind (Gagne & Coppola, 2017) and spatial reasoning (Gentner et al., 2013). People who suffer language impairments in adulthood (e.g., stroke-induced aphasia) show impairments in putatively "nonverbal" (fluid) reasoning (Baldo et al., 2015), and in tasks requiring selectively attending to a particular dimension, e.g., appreciating that cherries and bricks are similar by virtue of their color (Koemeda-Lutz et al., 1987)—an ability that underlies much of abstract reasoning. Interestingly, they do not show pronounced deficits in theory of mind (Siegal & Varley, 2006), suggesting that the role of language may be to inform the initial development of theory of mind (after all, the major way we come to know what others are thinking is by having them tell us!). Experimental studies that manipulate language demonstrate its causal role in human cognition. For example, holding nonlinguistic experience constant, named categories, and categories consisting of more nameable parts are easier to learn (e.g., Zettersten & Lupyan, 2020); hearing a word activates more categorical mental representations (Edmiston & Lupyan, 2015) which is important for making inferences and for enabling compositional thought. Conversely, interfering with language impairs people's ability to learn rules and attend to specific dimensions (Lupyan, 2009), mirroring some of the impairments observed in aphasia.

**Neural dissociations do not imply lack of causality**

How can we square these findings with apparent neural dissociations between linguistic and nonlinguistic processes? One answer is that many aspects of language such as phonological and syntactic processing are highly specialized. As we gain linguistic expertise, these processes become modularized and neurally dissociable. But as the evidence above suggests, such emerging modularity does not mean that "nonlinguistic" cognition is independent of language.

To take one example of such interaction, although visual processing is by no means a linguistic process, language actively modulates and constrains it. Language activates both primary visual cortex (Seydell-Greenwald et al., 2023) and higher-level visual regions (Ardila et al., 2015), and modulates basic visual processes. For example, a simple word can make otherwise invisible objects visible (Lupyan & Ward, 2013). The addition of language input also vastly improves the match between visual representations in people and those learned by artificial neural networks trained on images (Wang et al., 2023; see also Bi, 2021). Despite being dissociable, language informs and shapes visual processing. These causal links do not imply that the medium of thinking or perceiving is linguistic. Rather, they suggest that forming useful internal models (both developmentally, and when executing a task) benefits from learning and using the abstractions provided to us by natural language.

Investigations of how LLMs learn to process language (e.g., noun-verb agreement) are revealing the emergence of specialized circuits (Tigges et al., 2024). But this does not license distinguishing "formal" linguistic competence that underlies LLMs' ability to produce fluent coherent language from "functional linguistic" competence which involves using language for downstream tasks (see Mahowald et al., 2024 in recommended readings). It is indeed useful to distinguish linguistic tasks like noun-verb agreement from various "functional" downstream tasks like medical diagnosis. LLMs can do both, but the mechanisms are likely quite different because the computational needs of the tasks are different. Yet it would be clearly a mistake to conclude from such dissociations that the ability of LLMs to diagnose patients is independent of language. It is of interest that the performance of many (though importantly, not all) downstream tasks can be improved not by augmenting training with nonlinguistic materials, but simply by having models use internal language—analogous to inner speech—before producing a response (Chen et al., 2025).

**Why does language have these effects?**

Why does language have these effects on people and how can exposure to natural language, no matter its scale, enable LLMs to perform so many different types of cognitive tasks? One reason is that language provides a set of abstractions encoded into its vocabulary. While it may seem that words simply reflect natural categories (the "joints of nature"), in fact vocabularies are products of collective intelligence. Very few if any of us are capable of inventing number words to denote cardinalities, but once these words exist, we readily learn them in the course of learning the rest of language. Absent number words, we struggle to represent exact numerosities even when living in a numerate culture (Spaepen et al., 2011). The vocabulary of every language consists of thousands of such pre-discovered abstractions that are much easier to learn than to reinvent. Vocabularies and larger verbal constructions can thus be viewed as highly generative compression schemes for capturing the human *Umwelt* (see also Rothschild, forthcoming; Clatterbuck & Gentry, 2025).

When we expose sufficiently powerful statistical learning systems (such as transformers) to language and force them to get good at predicting the next word, they come to learn not just the statistical patterns of language, but generative model of the latent structure that produced the language, that is, the world through human eyes, including perceptual relationships (Marjieh et al., 2024) and causal links (Kıcıman et al., 2024) thought to be unlearnable from passive observation, much less from passive observation of language (Sloman, 2005). To paraphrase a [recent social media post by @gracekind.net](): What does ChatGPT do? It predicts the next token. What does it do to predict the next token? Whatever it takes.

**Human and machine intelligence as collective, language-based intelligence**

In trying to understand the astonishing intellectual achievements of our species, it has been common to appeal to the computational power of individual minds. Recognizing that language is both the product of our collective intelligence and also its shaper, encourages a more humble view: the power of human intelligence has less to do with our individual mental firepower and more to do with the scaffolding provided by language (and other aspects of culture, Henrich, 2015). The presence of certain words in a vocabulary ensures that all the members of the language community learn the categories these words denote. Learning these categories paves the way for their creative recombination. On this view, language is not just a communication medium; its abstractions help us form internal models we use to cognize the world.

Adopting this perspective makes the successes of artificial neural networks trained on language less surprising: the open-ended nature of language means that it can be used to convey everything from how we feel, to recipes, to scientific findings. Reducing prediction error across these varied domains turns out to be a highly effective means to learn the latent structure of these data: the human *Umwelt*. Despite the vast differences between LLMs and human minds, language appears to help both. As our creations, the intelligence of LLMs depends on the cognitive labor of countless human minds. So too does our own intelligence.

**Recommended readings:**

- Aguera y Arcas, B. (2025). *What Is Intelligence?: Lessons from AI About Evolution, Computing, and Minds*. MIT Press. https://mitpress.mit.edu/9780262049955/what-is-intelligence/
  A wide-ranging exploration of prediction as a central process that biological and artificial systems use to build models of themselves and the world.

- Piantadosi, S. T. (2024). Modern language models refute Chomsky's approach to language. *From Fieldwork to Linguistic Theory*, 353–414.

  This position piece provides a comprehensive critique of traditional generative linguistics based on the successes of large language models in learning and using language.

- Wang, Z., Akshi, Keil, S., Kim, J. S., & Bedny, M. (in press). Constructing meaning from language: Visual knowledge in people born blind and in large language models. *Annual Review of Linguistics*.

  A comprehensive review of the literature on the knowledge of visual appearance in people born blind and large language models, drawing parallels on the aspects of the visual world that are, in principle, learnable from language.

- Mahowald, K., Ivanova, A. A., Blank, I. A., Kanwisher, N., Tenenbaum, J. B., & Fedorenko, E. (2024). Dissociating language and thought in large language models. *Trends in Cognitive Sciences*, *28*(6), 517–540. https://doi.org/10.1016/j.tics.2024.01.011

  A provocative review arguing for a dissociation between formal and functional linguistic competence in LLMs. Although we think the performance profile of these models fails to support this distinction, it is useful to understand the different demands that language production and "thinking" place on a system.